\documentclass{article} %
\usepackage{uxbench}

\usepackage{booktabs}
\usepackage{graphicx}
\usepackage{enumitem}
\usepackage{wrapfig}
\usepackage{algorithm}
\usepackage{algpseudocode}
\usepackage{wrapfig}
\usepackage{float}
\usepackage{microtype}
\usepackage{amsmath}
\usepackage{amssymb}
\usepackage{colortbl}
\usepackage[utf8]{inputenc}
\definecolor{lightgray}{rgb}{0.9,0.9,0.9}
\usepackage{caption}
\usepackage{subcaption}
\usepackage{xcolor}
\usepackage{setspace}
\usepackage{url}
\usepackage{multirow}
\usepackage{colortbl}
\usepackage{tabularx}
\usepackage{blindtext}
\usepackage{pgfplots}
\pgfplotsset{compat=1.18} 
\usepackage{tikz}
\usetikzlibrary{er,positioning,bayesnet}
\usepackage{makecell}
\usepackage{tipa}
\usepackage{siunitx}
\usepackage{nicefrac}
\usepackage{tocloft}
\usepackage{listings}
\usepackage[raster,skins]{tcolorbox} %
\usepackage{xltabular}
\usepackage{adjustbox}
\usepackage{xurl}

\definecolor{ybdeep}{HTML}{12A594}
\definecolor{ybrule}{HTML}{0B7A6B}
\definecolor{ybgreen}{HTML}{0E9E86}
\definecolor{ybtint}{HTML}{E6F5F1}
\definecolor{ybtintb}{HTML}{F3FBF9}
\definecolor{ybneg}{HTML}{B4442F}
\definecolor{ybgrey}{HTML}{8A8F98}

\makeatother

\usepackage{graphicx}

\usepackage{amsmath, amssymb}
\usepackage{booktabs}
\usepackage{multirow}

\usepackage{algorithm}
\usepackage{algpseudocode}
\usepackage{pifont} 

\usepackage{bbm}
\usepackage{tikz}
\newlength{\barwidth}
\usepackage{lipsum}

\usepackage{hyperref}
\usepackage{CJKutf8}

\title{UXBench Pro: Benchmarking Personalized User Experience in Multi-Turn Dialogue Interactions}

\author{
Mengze Hong \quad Zeyang Lei\quad Wenbo Shang \quad Xia Zeng \quad Xiying Zhao \\ Qi Zhu \quad Chen Jason Zhang \quad Di Jiang \quad Taiming Fu \quad Qiongyi Zhou \quad Qinghe Chang \quad Fubao Zhang \quad Chenxuan Ma \quad Minlong Peng \quad Jinfeng Huang \quad Zineng Zhou \quad Jindou Wu \quad Muge Qi \quad Sijun He \quad Xin Cui \\ Di Liang \quad Yuan Hua \quad Davey Chen\thanks{Project Leader: Davey Chen}
\\ \vspace{0.7em}
Hong Kong Polytechnic University \qquad Yuanbao Team, Tencent
}

\begin{document}

\maketitle

\begin{abstract}
Evaluating user experience (UX) with automated computational methods has gained increasing attention, supported by empirical evidence from UXBench. However, binary preference prediction provides limited insight, while relying on a single user-agnostic reward model overlooks the inherent heterogeneity of users, whose expectations can differ substantially. In this paper, we present UXBench Pro, comprising 1{,}000 test instances derived from real user interactions across 12 task scenarios and 82 domains. Each instance is paired with a FACTORS user profile that characterizes the user through seven interpretable behavioral facets, differentiating user groups. To provide richer evaluation insights, we introduce a dual-perspective paradigm that combines a personalized User Reward Model (URM) for third-person judgment with Sim4Eval, a user simulator that enables multi-turn interactions and provides first-person evaluation across four cognitive state dimensions. To assess the reliability of these based evaluators, we further introduce two meta-benchmarks, URMBench and USimBench, that evaluate how faithfully they reproduce real human preferences and behaviors. Extensive experiments reveal seven key findings that highlight the importance of user modeling and multi-perspective evaluation, offering a fresh perspective on user-centric benchmarking and motivating personalized model optimization.
\end{abstract}
\maketitle

\begin{figure}[!h]
    \centering
    \includegraphics[width=\linewidth]{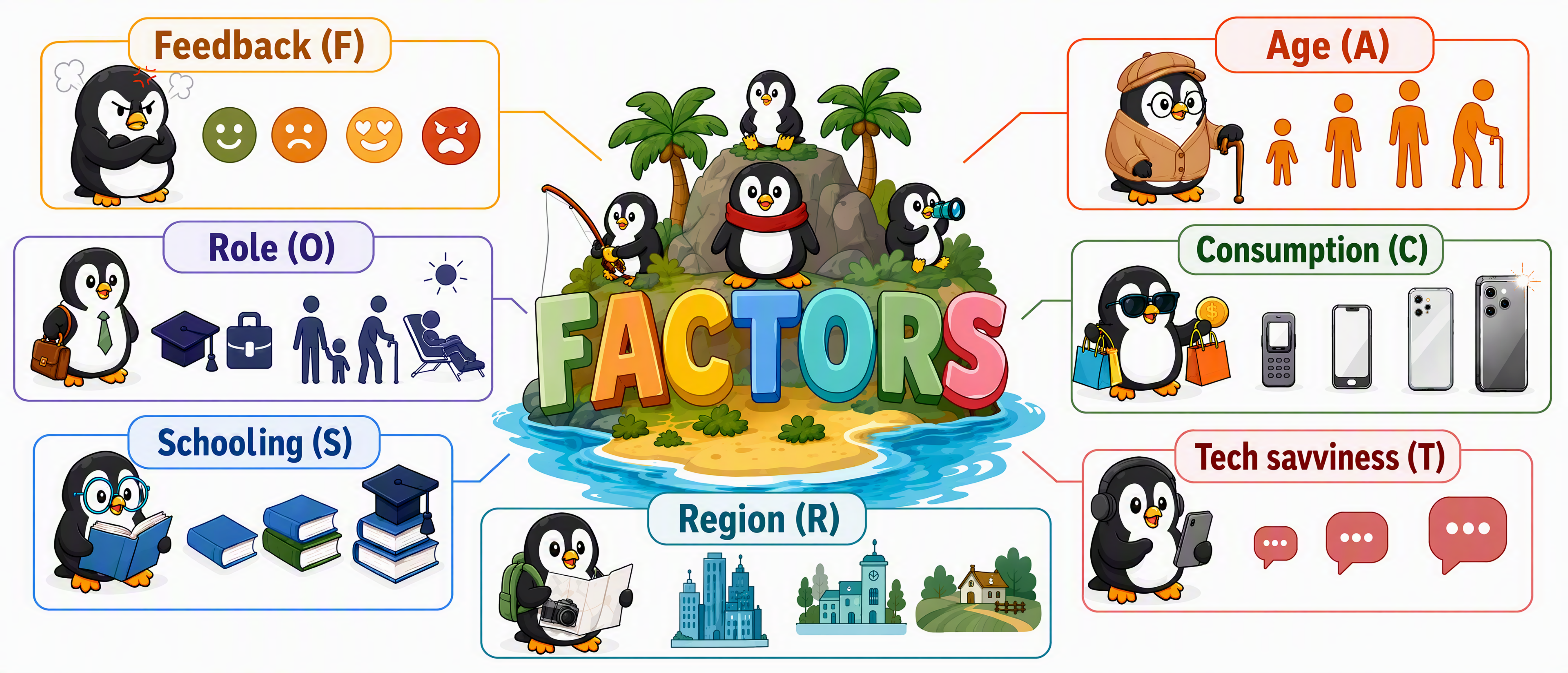}
    \caption{Illustration of the seven-dimensional \emph{FACTORS} profile schema for characterizing heterogeneous users in human-AI interactions.}
    \label{fig:factors}
\end{figure}

\section{Introduction}

As AI assistants engage with billions of users, defining what makes a model ``good'' becomes increasingly context-dependent. Existing benchmarks often prioritize challenging problems that test model capabilities and expertise \citep{yang2026onemillionbenchfarlanguageagents, phan2025humanity, hong-etal-2025-qualbench}. While such evaluations provide meaningful signals for model development, they offer limited insight into end-user experience, as performance on difficult edge cases may not reflect how users perceive and engage with AI assistants in everyday interactions. User experience is a complex interplay of model capability, linguistic style, and the user's expectations and cognitive state, making the current evaluation landscape a limited proxy.
\textbf{UXBench} \citep{uxbench2026} represents an initial step toward the vision of user-centric benchmarking, demonstrating that user experience can be quantified via computational rewards calibrated to real human feedback signals, elevating UX from an abstract, human-centered concept to a measurable evaluation objective. However, several limitations in existing work motivate further exploration in user-centric benchmarking:

\begin{enumerate}
\item Binary user feedback (like/dislike), while being straightforward to calibrate to real user behavior, provides coarse-grained signals with limited interpretability and offers little insight into the underlying factors that drive satisfaction in human-AI interactions.

\item One user's satisfaction does not necessarily imply another's. A single user-agnostic judge model trained on aggregated user feedback implicitly assumes a universal notion of user experience, overlooking the heterogeneity in individual preferences, behavioral tendencies, and interaction contexts.

\item Existing evaluations primarily focus on isolated single-turn interactions, overlooking the dynamic evolution of user experience across multi-turn conversations, where satisfaction may accumulate over time.
\end{enumerate}

Users are inherently heterogeneous in age, educational background, professional expertise, and expectations of what constitutes a ``good'' response, and this heterogeneity inevitably extends to their expectations. A retired senior seeking medication advice, a software engineer debugging a compiler error, and a teenager requesting creative writing support may all interact with the same AI system, yet their criteria for satisfaction can differ substantially. Similarly, a verbose, information-dense response that one user perceives as unnecessary may be exactly what another values for its completeness. Intuitively, general-purpose AI needs to serve heterogeneous users along two dimensions: users ask different queries, and different users perceive the same response differently. Building personalized AI assistants is important for supporting diverse user populations \citep{yuan-etal-2025-personalized} and should therefore be reflected in evaluation frameworks. When user experience is modeled as user-invariant, individually valid preferences are collapsed into a single aggregate signal, obscuring meaningful differences in what users value.

In this paper, we take a step toward automated computational methods that enable personalized benchmarking for human-AI interactions, modeling user experience as jointly conditioned on \emph{who} the user is and \emph{what} they ask \citep{Hassenzahl01032006}. We propose \textbf{UXBench Pro}, a benchmark comprising 1{,}000 high-quality user queries, each paired with a user profile. Guided by the hypothesis that offline benchmarking should approximate online user distributions \citep{nie2026characterflywheelscalingiterativeimprovement}, we design the \textbf{FACTORS} user profile schema to characterize users across seven behavioral facets (see Figure \ref{fig:factors}). To enable fine-grained and informative assessment, we introduce a dual-perspective paradigm that combines a User Reward Model (URM) with a user simulator, supporting both third-person inspection and first-person assessment of model responses in multi-turn interactions. Finally, we propose the \textbf{meta-benchmarking principle} with URM-Bench and USimBench to sufficiently validate model-based evaluators against human values to ensure that benchmark outcomes reflect calibrated signals of human preference and behaviors.

The contributions of this work span both methodological innovations and empirical insights:

\begin{enumerate}
\item \textbf{Personalized UX Evaluation.}
We release UXBench Pro, a user-centric benchmark incorporating a seven-facet \textbf{FACTORS} user profile schema and covering 12 task scenarios across work and life use cases. Our experiments reveal systematic effects of user heterogeneity on personalization and UX evaluation, yielding seven key findings that support user-centric benchmarking and AI assistant optimization.

\item \textbf{Dual-Perspective Evaluation.}
We introduce a user reward model that serves as a personalized third-person inspector, accompanied by Sim4Eval, a user simulator that supports multi-turn dialogue and offers first-person evaluation by outputting user states across four cognitive state dimensions, providing interpretable evaluation insights.

\item \textbf{Meta-benchmarking Principle.} We propose the meta-benchmark principle and introduce two useful benchmark resources: URM-Bench and USimBench, which use real user preferences and interaction behavior as ground truth signals to evaluate judge models before their deployment in benchmarking AI assistants.
\end{enumerate}

This work is intended to contribute primarily at the conceptual and methodological level, rather than by emphasizing the relative performance of particular models. As LLMs continue to evolve rapidly, model-specific comparisons can quickly become outdated with successive generations. We therefore focus on developing generalizable principles for understanding and evaluating user experience, moving beyond static and user-agnostic notions of model quality toward evaluation that accounts for the heterogeneous and evolving nature of human-AI interaction. We hope these principles and findings remain useful for evaluating future generations of AI assistants.

\section{Related Work}
\label{sec:related_work}

\subsection{User Experience Evaluation} 

User experience (UX) has long been a central concern in HCI and is traditionally evaluated through direct user involvement, including usability studies, surveys, interviews, focus groups, and field studies~\citep{dumas2006usability}. Standardized instruments such as AttrakDiff, UEQ, and meCUE quantify subjective experience and form the basis for rubric-based judgments~\citep{diazoreiro2021ux}. These methods provide rich and interpretable feedback, but they are costly to scale: a systematic review of 548 UX studies reports a median of only 20 participants per study~\citep{diazoreiro2021ux}, introducing bias against underrepresented user groups. This has motivated computational approaches that augment or automate parts of UX evaluation~\citep{ivory2001state}, from analyzing interaction traces and behavioral signals to recent LLM-based methods that simulate users and generate usability feedback~\citep{xiang2024simuser}.

The scalability challenge is particularly pronounced for conversational AI assistants. Most existing LLM benchmarks focus on capabilities such as reasoning, knowledge, and instruction-following, offering limited insight into the quality of the experience users perceive. Prior work has modeled user satisfaction from graded dialogue annotations~\citep{sun2021uss} or implicit behavioral signals such as interruptions and rephrasing~\citep{park2021scalable}, but these approaches are largely confined to task-oriented corpora and predefined rating schemes. UXBench~\citep{uxbench2026} takes an important step toward grounding UX evaluation in real user feedback, showing that frontier LLMs systematically overestimate interaction quality, whereas a feedback-trained reward model predicts user satisfaction more reliably. However, UXBench remains a static, single-turn benchmark that models users from a generic perspective, leaving open how UX signals evolve and accumulate across multi-turn interactions and heterogeneous individual users.

\subsection{Reward Modeling}

Reward models (RMs) are central to aligning LLMs with human preferences, serving both as reward signals in RLHF and as scorers for inference-time selection~\citep{ouyang2022instructgpt}. Recent approaches emphasize direct preference optimization and leverage generative reward models (GRMs) to evaluate responses through natural-language reasoning. As RMs become more widely used, directly evaluating their quality has become increasingly important. RMB evaluates reward models across $49$ real-world scenarios using a Best-of-$N$ protocol and shows that standard pairwise accuracy correlates only weakly with downstream alignment quality, while generative RMs appear particularly promising~\citep{zhou2025rmb}. RewardBench 2 further addresses contamination and ceiling effects by constructing a harder benchmark from fresh human prompts~\citep{malik2025rewardbench2}. However, these benchmarks are still built from curated preference data and mainly measure correctness or alignment utility, rather than whether an RM captures the experience and preferences of different users, motivating a user-centric RM and benchmark.

\subsection{User Simulation for Multi-turn Dialogues}

Early user simulators learn user policies from corpora, e.g., sequence-to-sequence generators for RL-based dialog systems~\citep{shi2019build} and data-driven metaphorical testers~\citep{sun2023metaphorical}. With large language models, the paradigm has largely shifted to LLM-based simulation along three lines. One line uses the LLM as a generic user proxy via in-context learning or prompting~\citep{terragni2023incontext,luo2024duetsim}. Another augments it with goal-state tracking or verification for goal consistency across turns~\citep{mehri2026goal}. A third adopts a persona- or digital-twin view that models user differences through personas, behavior chains, or implicit profiles~\citep{wang2025usp,xie2025human,li2025behaviorchain,du2026twinvoice}. Despite gains in fluency and controllability, two limitations remain central to our study. First, these methods are largely domain-specific, tightly coupled to particular tasks, ontologies, or benchmarks such as MultiWOZ~\citep{eric2020multiwoz}, and thus transfer poorly to open-domain, assistant-style interaction. Second, their user-profile modeling is weak: profiles are typically hand-authored, or coarse-grained prompt prefixes rather than stable, fine-grained signals inferred from real interactions.

\begin{figure}[!t]
    \centering
    \includegraphics[width=1\linewidth]{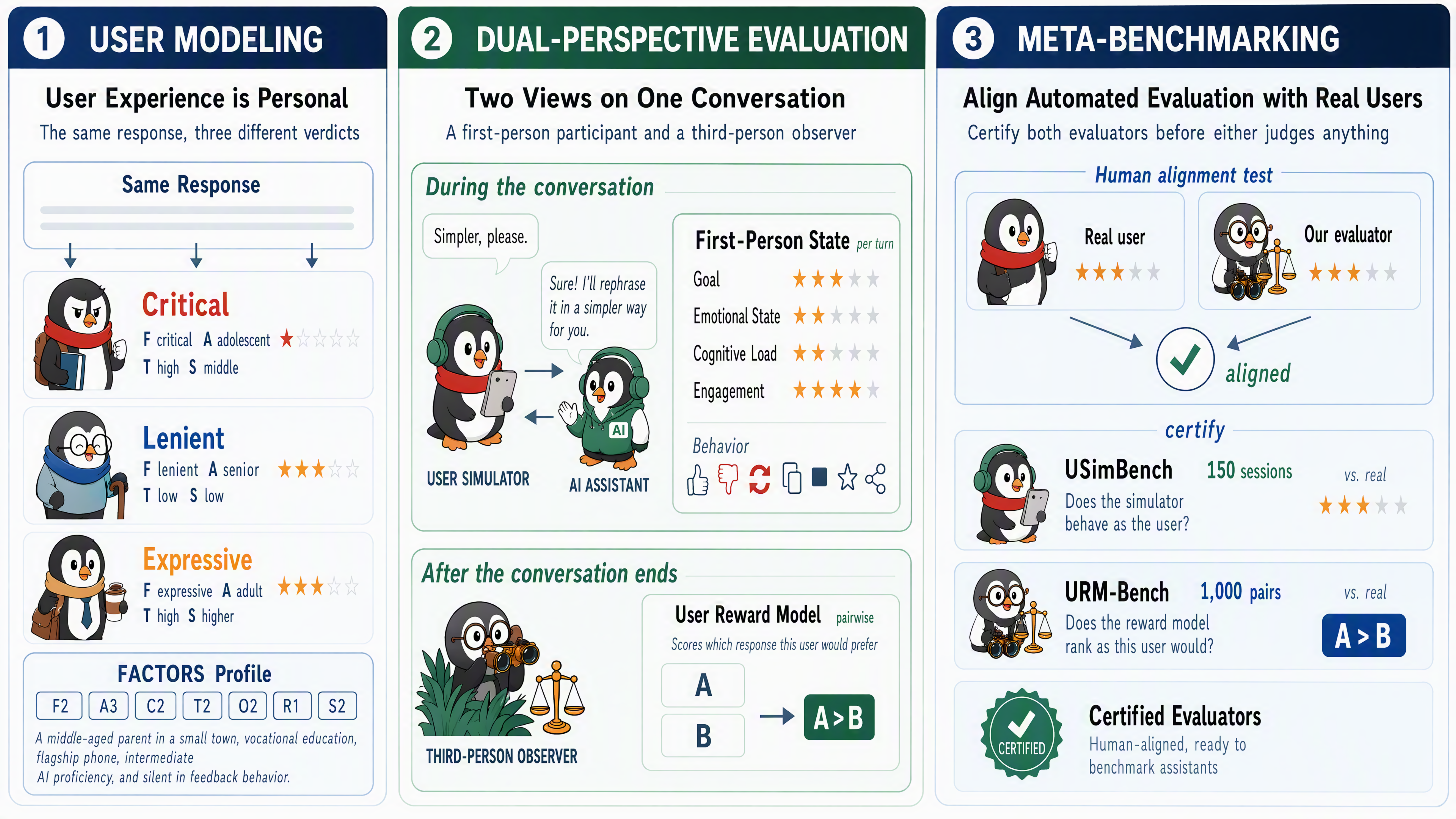}
    \caption{Overview of UXBench Pro with three core proposed methodologies.}
    \label{fig:overview}
\end{figure}

\section{UXBench Pro}
\label{sec:uxbench_pro}

\subsection{Overview}

UXBench Pro is a benchmark for evaluating personalized user experience, comprising 1,000 high-quality test instances derived from interaction logs of a mainstream AI assistant in China that serves a demographically diverse user population. It is designed to answer three core research questions: (1) whether modeling user characteristics can improve the calibration of automated UX judgment; (2) whether injected user knowledge can improve the response quality that better satisfies heterogeneous users; and (3) whether user-aware evaluation provides more human-aligned insights for model improvement. We hypothesize that user experience is conditioned by both the user profile and the task context, recognizing that the same response may be perceived differently by different users. 


To support reliable personalized evaluation, UXBench Pro treats the evaluator itself as an object of evaluation. We therefore introduce two complementary meta-benchmarks, \textbf{URM-Bench} (Section \ref{sec:urm_bench}) and \textbf{USimBench} (Section \ref{sec:usimbench}), which use real user feedback and interaction behavior to assess whether the URM and user simulator reproduce human-aligned outcomes. By comparing model performance with and without user profiles, these benchmarks further isolate the contribution of personalization, enabling a more reliable analysis of how user profiles shape experience.

\begin{table}[t]
\centering 
\small
\setlength{\tabcolsep}{4pt}
\caption{\textbf{FACTORS} user profile schema, with each dimension discretized into interpretable categories plus an unknown state for completeness.}
\begin{tabular}{lll}
\toprule
\textbf{Dimension} & \textbf{Signal} & \textbf{Categories} \\
\midrule
Feedback ($F$) & Likes/dislikes, 30d & Silent, critical, lenient, expressive, light \\
Age ($A$) & Age interval & Adolescent, young adult, adult, senior, unknown \\
Consumption ($C$) & Device tier & Low, medium, high, flagship, unknown \\
Technical Usage ($T$) & Prompt volume + usage & Light, regular, heavy \\
Role ($O$) & Dialogue-mined life stage & Student, working-age, retired, parenting, unknown \\
Region ($R$) & Geographic location & Metro, town, rural, unknown \\
Schooling ($S$) & Mined education signals & Low, middle, higher, unknown \\
\bottomrule
\end{tabular}
\label{tab:factors_profile}
\end{table}

\subsection{User Profile Schema with \emph{FACTORS}}
\label{sec:schema}

Modeling user heterogeneity is challenging. UXBench Pro introduces an explicit, interpretable user profile schema with seven dimensions derived from behavioral and contextual signals, organized as the \emph{\textbf{FACTORS}} schema (Table~\ref{tab:factors_profile}). Each dimension is discretized into meaningful categories with an additional ``unknown'' state (Appendix~\ref{appendix:profile_encoding}). Profiles are derived from sanitized dialogue logs or device-level information and are intended as contextual signals for personalization rather than definitive user characterizations, minimizing privacy risks and minimizing exposure of identifying information. Each user is represented by a seven-component profile code, which is rendered into a human-readable persona card for conditioning the personalized reward model and user simulator. The example below illustrates the profile code, facet-level decoding, and persona card.

\begin{tcolorbox}[colback=ybtintb,colframe=ybdeep!50,boxrule=0.4pt,arc=1.2pt,left=5pt,right=5pt,top=3pt,bottom=3pt,fontupper=\small]
\textbf{Profile code}\quad $F2\,{\cdot}\,A3\,{\cdot}\,C2\,{\cdot}\,T2\,{\cdot}\,O2\,{\cdot}\,R1\,{\cdot}\,S2$\\[2pt]
\textbf{Decoded}\quad critical feedback ($F2$), adult ($A3$), high-tier device ($C2$), heavy AI usage ($T2$), working-age ($O2$), metropolitan ($R1$), higher schooling ($S2$)\\[2pt]
\textbf{Persona card}\quad \emph{A working-age adult in a major city, well educated and a frequent user of AI tools, using a high-end device, who tends to be critical and dislike-prone when judging responses.}
\end{tcolorbox}

\begin{figure}[t]
\centering 
\includegraphics[width=1\linewidth]{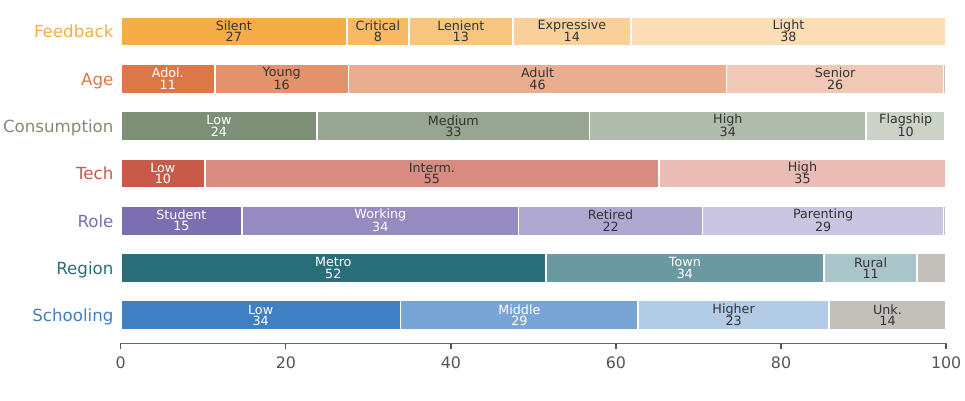}
\vspace{-2.2em}
\caption{Distribution of the 1{,}000 test cases across each FACTORS facet.}
\label{fig:factors_dist}
\end{figure}

To characterize the user population, Figure~\ref{fig:factors_dist} shows the empirical distributions of all seven FACTORS dimensions across the 1{,}000 instances. The benchmark spans diverse age, consumption, region, and education groups, while feedback disposition is dominated by silent and light users, reflecting the sparsity of explicit feedback. Critical, lenient, and expressive users are retained at informative proportions through the sampling procedure in Section~\ref{sec:grid}. Specifically, we derive the feedback disposition from the user's 30-day feedback history using the like count $L$, dislike count $U$, total feedback volume $T=L+U$, and positive feedback ratio $L/T$.

\subsection{User x Task Grid}
\label{sec:grid}

In the released dataset, each evaluation instance is represented as $(x,u,t)$, where $x$ denotes the interaction context, $u$ the FACTORS user profile, and $t$ the task scenario. Accordingly, we define the evaluation space as $\mathcal{G}=\mathcal{U}\times\mathcal{T}$, enabling systematic analysis of how model performance varies across scenarios, and how user profiles and task scenarios jointly shape user experience. The task axis follows a hierarchical taxonomy with two high-level categories, \emph{work} and \emph{life} \citep{chatterji2025people}, which are further divided into 12 fine-grained scenarios (see Table~\ref{tab:task_taxonomy}). Scenario labels are assigned at the session level via majority voting on dialogue signals mined using intent and domain classifiers.

To balance task coverage without distorting natural profile--task correlations, we balance only across task categories while preserving within-task profile distributions. For each task $t$, we sample
\[
|\mathcal{D}_t|=\min(n_t,K),
\]
retaining rare scenarios in full and capping dense ones at $K$. This yields comparable per-scenario counts (normalized task entropy: $0.953$), while FACTORS distributions emerge naturally within each task. We further limit repeated users, exclude sessions with fully unknown profiles, and impose a 10\% minimum representation floor for adolescent and rural users, as well as for lenient and expressive feedback dispositions.


\begin{table}[t]
\centering 
\caption{Task taxonomy and per-scenario instance counts in the released set.}
\begin{tabular}{lllr}
\toprule
\textbf{Category} & \textbf{Code} & \textbf{Scenario} & \textbf{$N$} \\
\midrule
\multirow{5}{*}{Work}
& W1 & Office writing (documents, emails, reports) & 107 \\
& W2 & Coding (programming, debugging) & 17 \\
& W3 & Data analysis (analytics, decision support) & 110 \\
& W4 & Professional lookup (legal, financial, medical) & 39 \\
& W5 & Translation (business translation) & 18 \\
\midrule
\multirow{7}{*}{Life}
& L1 & Study assistance (learning, homework) & 104 \\
& L2 & Everyday services (planning, recommendations) & 98 \\
& L3 & Health and wellness & 85 \\
& L4 & Emotional interaction (companionship, chit-chat) & 125 \\
& L5 & Entertainment (creative generation, leisure) & 98 \\
& L6 & Personal finance (budgeting, planning) & 95 \\
& L7 & General information (open-domain queries) & 104 \\
\bottomrule
\end{tabular}
\label{tab:task_taxonomy}
\end{table}

\subsection{Data Construction Pipeline}
\label{sec:pipeline}

\paragraph{Stage 1, data cleaning.}
Starting from 71{,}221 raw sessions with explicit negative feedback to ensure sufficient difficulty, we deduplicate at three granularities and remove degenerate interactions, including empty or overly short prompts and responses, punctuation- or emoji-only turns, speech-recognition failures, no-op requests, and garbled text.

\paragraph{Stage 2, safety review and de-identification.}
Each session undergoes two-level lexicon screening across seven risk categories: sexually explicit content, violence, political sensitivity, illegal activities, extremism, drug-related content, and self-harm. De-identification is applied at three layers. At the schema layer, released records contain no user identifiers, and all profile attributes are represented only as coarse buckets. At the text layer, a twelve-class rule-based scanner detects residual PII, including phone numbers, email and network addresses, physical addresses, and personal names. Finally, queries are fully rewritten by a small, locally hosted model (hunyuan-3) to remove any remaining identifying details while preserving the original intent.

\paragraph{Stage 3, multi-agent quality assurance.}
We use multi-agent debate to filter instances in which negative user feedback may stem from ambiguous or unanswerable queries rather than genuine model deficiencies. A proposer first argues why an instance should be retained, and three independent critics evaluate whether the query is meaningfully challenging, whether the assistant response is genuinely deficient, and whether the instance can discriminate between stronger and weaker systems. Retention follows a deterministic voting rule: a session is rejected if any critic issues a high-confidence refutation, if at least two critics refute the proposal, or if multiple critics challenge it without any supporting vote. This procedure retains 58.5\% of sessions, flags 7.6\% as low-value, rejects 32.7\%, and loses 1.1\% to technical failures. Balanced sampling is performed only after this stage to preserve the intended user--task distribution.

\paragraph{Stage 4, human review.}
Finally, ten annotators conduct a systematic review under a ring-based double-coverage scheme, so that every session receives two independent reviews while each annotator covers two adjacent strata. Annotators see only the dialogue and candidate responses, without internal labels or user-profile information. They assess content safety, identify privacy issues, and, for the URM-Bench subset (Section~\ref{sec:urm_bench}), indicate their preferred response. The protocol favors recall by requiring uncertain cases to be flagged for adjudication. Content-safety judgments achieve 0.971 raw agreement, and any instance flagged by at least one annotator for safety or privacy concerns is excluded from the released dataset.

\subsection{Data Statistics}
\label{sec:data_stats}

\begin{figure}[t]
\centering
\includegraphics[width=0.9\linewidth]{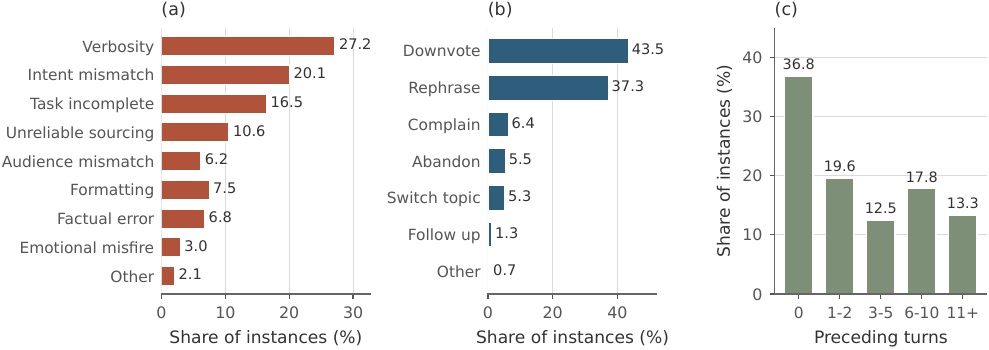}
\caption{Composition of the experience signals: (a) failure dimension, (b) user actions, (c) preceding dialogue turns.}
\label{fig:uxbench_stats}
\end{figure}

Table~\ref{tab:bench_stats} summarizes the benchmark statistics, and Figure~\ref{fig:uxbench_stats} shows the composition of its experience signals. The benchmark spans 12 task scenarios and 82 fine-grained domains, with 63.2\% of instances containing preceding dialogue context and 85.0\% having all seven FACTORS facets resolved. Verbosity, intent mismatch, and task incompleteness are the most common failure dimensions, jointly accounting for 63.8\% of instances, whereas factual errors account for only 6.8\%. The benchmark also captures diverse dissatisfaction signals: 11.5\% of instances contain explicit verbal complaints, while 39.7\% involve user recovery attempts. Interactions contain an average of 4.7 preceding turns, with broad coverage across tasks, user profiles, and interactions, while preserving realistic multi-turn settings.

\begin{table}[t]
\centering
\caption{Summary statistics of UXBench Pro.}
\begin{tabular}{lr}
\toprule
\textbf{Statistic} & \textbf{Value} \\
\midrule
Test instances & 1{,}000 \\
Task scenarios & 12 \\
Domains (coarse / fine) & 15 / 82 \\
Work / life split & 29.1\% / 70.9\% \\
Normalized task entropy & 0.953 \\
Instances with a preceding dialogue context & 63.2\% \\
Preceding turns (mean / median) & 4.7 / 2 \\
Response length in characters (median / p90) & 738 / 1,786 \\
Profiles with all seven facets resolved & 85.0\% \\
\bottomrule
\end{tabular}
\label{tab:bench_stats}
\end{table}

\section{Methodology}
\label{sec:method}

We propose a dual-perspective framework in which two complementary evaluators assess an assistant's response. The third-person perspective is provided by a \textbf{User Reward Model} (URM), which estimates response satisfaction conditioned on the user's profile. The first-person perspective is provided by a \textbf{user simulator} via \textbf{Sim4Eval}, which continues the interaction and exposes interpretable cognitive state dimensions associated with the evolving UX.

\subsection{User Reward Model}
\label{sec:urm}

The URM evaluates a candidate assistant's response given an authentic dialogue context and the user's FACTORS profile, and an effective URM is key to reliable UX judgment. We consider reward models sharing a common base model (Hunyuan-3) but differ along two axes: \emph{modeling formulation} and \emph{preference signal}. For formulation, we compare three variants: a pointwise GRM that scores a response from the normalized probabilities of positive and negative judgment tokens, a Bradley-Terry reward model (BTRM) with a classification head trained on preference pairs, and a pairwise GRM that retains the generative formulation while directly modeling paired comparisons between two responses. For preference signal, we consider three sources: regeneration preference, where two responses to the same context are compared with the assumption that the regenerated (later) response better satisfies the user; dual-answer preference, where users select between two candidate responses; and pointwise feedback, where individual responses are labeled as liked or disliked, following a similar design in UXBench.

\subsection{Sim4Eval Framework}
\label{sec:usim}

The URM evaluation cannot capture how user experience accumulates and varies across multiple turns. Thus, we introduce a Sim4Eval approach that adopts the first-person perspective of a user simulator: at each turn $t$, it produces the next user query $v_t$, a four-dimensional cognitive state $z_t\in\{1,\ldots,5\}^4$, and a continuation decision $c_t\in\{\textsc{continue},\textsc{leave}\}$, all conditioned on the latest assistant response $a_t$ and the injected user profile. The four state dimensions are goal progress, emotional fit, cognitive fit, and engagement preference (see detailed breakdown in Appendix \ref{appendix:state_rubric}). These dimensions reflect complementary factors associated with user experience: progress toward task goals and interaction costs~\citep{walker1997paradise}, affective and empathic alignment~\citep{liu2018machines}, cognitive effort during human--agent interaction~\citep{daronnat2021inferring}, and users' heterogeneous preferences for conversational style and engagement~\citep{bhattacharjee2024understanding}.

To select a simulator that best matches real user queries and interaction behavior, we consider naive prompting, supervised finetuning on 21,313 authentic sessions, and a proposed RAG-based simulator. A user profile specifies who a user is, but not how that user tends to interact with AI. We therefore introduce the \textbf{Retrieval-Augmented User Simulator}, built on $N=135{,}915$ authentic follow-up turns from roughly $36{,}000$ sessions, $\mathcal{D}=\{(\tilde a_i,\tilde v_i)\}_{i=1}^{N}$. Using separate TF-IDF encoders $\Phi_a$ and $\Phi_u$ for assistant responses and user utterances, we score each candidate as
\begin{equation}
s_i =
\alpha\cos\left(\Phi_a(a_t),\Phi_a(\tilde a_i)\right)
+
\beta\cos\left(\Phi_u(q_t),\Phi_u(\tilde v_i)\right),
\qquad
q_t=v_1\oplus v_{t-1},
\label{eq:retrieval_score}
\end{equation}
with $\alpha=0.7$ and $\beta=0.3$. The first term matches similar assistant responses, while the second captures conversational context from the opening and most recent user utterances. We retrieve the top 12 candidates, remove near-duplicates using character-level Jaccard similarity with threshold $\tau=0.5$, and retain $k=3$ diverse utterances. These raw utterances provide the simulator with examples of realistic user tone, length, and reaction style.

\subsection{Decomposing User Satisfaction}
\label{sec:decomposition}

We conceptually distinguish three sources of variation in conversational experience: response capability, interaction-state alignment, and profile-dependent preference. Response capability captures user-independent qualities such as factual correctness, task completion, and instruction compliance. Interaction-state alignment reflects how well a response fits the user's evolving goals, emotions, cognitive needs, and engagement preferences, while profile-dependent preference captures stable differences in how users value or tolerate response characteristics. The URM jointly assesses these factors from a third-person perspective, whereas Sim4Eval makes interaction-state alignment and profile-dependent reactions observable over multiple turns through simulated user states and behavior. This decomposition provides a conceptual lens for interpreting the complementary signals produced by the two evaluators, rather than assuming that user satisfaction can be reduced to a single universal notion of response quality.

\section{Meta-Benchmarking}
\label{sec:meta}

Meta-benchmarking is defined as the practice of evaluating an automated evaluator against real human behavior before deployment in a benchmarking system, treating the evaluator itself as the object of study rather than assuming alignment with human behavior. Without such validation, systematic divergence from human preferences can propagate into downstream biases in evaluation results. Aligned with the evaluation framework, we construct two meta-benchmarks from the interaction logs, using the same data-cleaning and verification pipeline described in Section \ref{sec:pipeline}.

\subsection{URM-Bench}
\label{sec:urm_bench}

URM-Bench evaluates whether a reward model ranks responses according to real user preferences and whether conditioning on a FACTORS profile improves over a population-level baseline. Behaviorally observed pairwise preferences are extracted from regeneration and dual-answer dialogue logs, where the user-selected option is treated as the reference preference. We categorize each pair into four causes of a superior response, aligned with the user simulator's state dimensions, where the response either successfully recognized the user's present state (S1), inferred their underlying need (S2), adapted to their preferred style (S3), and respected their long-term preference (S4). The released benchmark contains 800 human-verified pairwise instances, 200 per mode. We report pairwise accuracy (PairAcc), in which exact ties count as errors. To test the value of profile conditioning, we compare paired outcomes with an exact two-sided McNemar test~\citep{mcnemar1947note}. Presentation order is fixed with the preferred response placed on either side near evenly, and the human reference is the mean PairAcc of ten contracted annotators on the same task.

\subsection{USimBench}
\label{sec:usimbench}

USimBench evaluates whether a user simulator produces multi-turn continuations faithful to real user behavior in feedback signals, semantic content, and interaction style. Each of the 150 instances is drawn from a real interaction session and includes a conversation prefix, the user's FACTORS profile, and a designated takeover point; the ground truth is the user's authentic continuation. Single-turn sessions are excluded, while feedback signal, conversation length, and takeover position (early, middle, or late) are balanced. The simulator generates per-turn utterances $v_t$, meta-behavioral signals $e_t$, self-reported core states $z_t$, and a termination decision $c_t$. We compare simulated and real continuations along three complementary
dimensions: feedback-signal fidelity,
semantic similarity of the generated continuation, and behavioral
similarity in interaction style. These metrics are reported separately
rather than collapsed into a single score, since linguistic resemblance
does not necessarily imply faithful user behavior.

\section{Results and Discussion}
\label{sec:results}

\subsection{UXBench Pro Results}
\label{sec:res_uxbench}

We use the best-performing URM and user simulator identified on the corresponding meta-benchmarks. Every model is evaluated on the same 1,000 instances under two controlled profile settings. First, we vary whether the user's FACTORS profile is provided to the assistant at generation time; the two prompts are otherwise identical, so any performance difference can be attributed to profile conditioning. Second, we independently vary whether the evaluator sees the same profile, yielding the four settings analyzed in Section~\ref{sec:res_profile}. In a pilot study, we observe a small average position bias ($+0.028$), although reversing candidate order changes the verdict in 9.3\% of comparisons. We therefore evaluate each comparison in both candidate orders and average the two results as $(p_{ab}+1-p_{ba})/2$.

\begin{table}[t]
\centering 
\small
\caption{UXBench Pro results for seven assistants. Turn BT and Traj. BT are Bradley--Terry strengths derived from pairwise URM judgments. Res.\ and Aban.\ denote resolution and abandonment rates. Separability is the fraction of assistant pairs with non-overlapping 95\% bootstrap confidence intervals; $\rho$ denotes Spearman correlation.}
\setlength{\tabcolsep}{5.5pt}
\renewcommand{\arraystretch}{1.08}
\begin{tabular}{l cc cccc cc}
\toprule
& \multicolumn{2}{c}{\textbf{URM (Pairwise GRM)}} & \multicolumn{4}{c}{\textbf{User state}} & \multicolumn{2}{c}{\textbf{Trajectory outcome}} \\
\cmidrule(lr){2-3}\cmidrule(lr){4-7}\cmidrule(lr){8-9}
\textbf{Assistant} & \textbf{Turn BT$\uparrow$} & \textbf{Traj.\ BT$\uparrow$} & \textbf{S1$\uparrow$} & \textbf{S2$\uparrow$} & \textbf{S3$\uparrow$} & \textbf{S4$\uparrow$} & \textbf{Res.$\uparrow$} & \textbf{Aban.$\downarrow$} \\
\midrule
Qwen3.7-Max      & $\mathbf{+0.170}$ & $\mathbf{+0.096}$ & 3.18 & 3.18 & 3.75 & 3.39 & 37.5 & \textbf{1.6} \\
DeepSeek-v4-pro  & $+0.047$ & $+0.053$ & \textbf{3.24} & 3.22 & 3.84 & 3.44 & \textbf{41.4} & 1.9 \\
Hunyuan-3        & $+0.028$ & $-0.027$ & 3.19 & \textbf{3.25} & \textbf{3.87} & \textbf{3.49} & 38.7 & 2.2 \\
GLM-5.2          & $+0.010$ & $-0.012$ & 3.15 & 3.20 & 3.79 & 3.39 & 34.5 & 2.1 \\
Claude-Opus-4.8  & $-0.014$ & $-0.006$ & 3.09 & 3.13 & 3.80 & 3.39 & 37.4 & 4.7 \\
GPT-5.5          & $-0.110$ & $-0.063$ & 3.14 & 3.13 & 3.79 & 3.37 & 34.0 & 2.2 \\
Gemini-3.1-Pro   & $-0.132$ & $-0.041$ & 2.94 & 3.01 & 3.71 & 3.23 & 30.7 & 1.9 \\
\midrule
\textbf{Separability$\uparrow$} & 100\% & 95.2\% & \multicolumn{6}{l}{Turn BT vs.\ Res.: $\rho=0.857$;\quad Turn BT vs.\ Traj.\ BT: $\rho=0.821$} \\
\bottomrule
\end{tabular}
\label{tab:ux_leaderboard}
\end{table}

\subsubsection{Overall Evaluation}
\label{sec:res_leaderboard}

\paragraph{\textbf{Finding 1: Third-person and first-person evaluation provide aligned but complementary views of assistant quality.}}
Table~\ref{tab:ux_leaderboard} compares response quality from the third-person URM perspective with interaction outcomes produced by Sim4Eval. The two views yield closely aligned rankings: Turn BT correlates strongly with resolution rate ($\rho=0.857$) and Traj.\ BT ($\rho=0.821$), while all assistant pairs are separable at the turn level and 95.2\% remain separable at the trajectory level. Qwen3.7-Max ranks highest in both Turn and Traj.\ BT, whereas Gemini-3.1-Pro has the lowest Turn BT and the lowest resolution rate (30.7\%). At the same time, the simulator exposes differences not captured by response ranking alone. Resolution and abandonment are nearly uncorrelated across assistants ($r=0.106$): Claude-Opus-4.8, for example, has a mid-range resolution rate of 37.4\% but the highest abandonment rate at 4.7\%. Abandoned trajectories win only 16.7\% of pairwise comparisons, compared with 47.6\% when neither trajectory is abandoned. Together, these results show that response quality, task resolution, and continuation behavior are related but non-equivalent aspects of conversational experience.

The simulator further helps characterize why interactions terminate. Across assistants, abandoned conversations consistently end with low goal progress: final goal-progress scores range from 1.32 to 1.69, whereas cognitive fit never falls below 3.06. For Claude-Opus-4.8, goal progress falls from 4.24 in satisfied exits to 1.55 in abandoned conversations, while cognitive fit remains relatively high at 3.53 versus 4.38. We also observe a strong association between routine offers to continue and early termination ($r=0.83$). Claude-Opus-4.8 ends 66.7\% of its first-turn responses with such an offer, compared with 35.2\% for the next-highest assistant and 21.4\% for the lowest, and records both the largest number of satisfied exits (152) and abandonments (47). Because this relationship may partly reflect the simulator's stopping policy, we treat it as a behavioral pattern rather than a causal effect; nevertheless, it illustrates how multi-turn evaluation can surface interaction dynamics invisible to response-level ranking.

\subsubsection{The Effect of User Profile}
\label{sec:res_profile}

\paragraph{\textbf{Finding 2: The apparent value of personalization depends strongly on who is evaluated and whether the evaluator knows the user.}}
Holding the response fixed, a profile-blind evaluator prefers the profile-free response in 62.5\% of comparisons, but this falls to 54.8\% when the evaluator sees the true user profile. Replacing the true profile with a scrambled one removes 57\% of this shift, indicating that the difference arises from user--profile matching rather than the mere presence of profile text. The effect is also highly heterogeneous across users: conditioned responses are preferred by expressive (62.8\%) and critical users (61.3\%), but not by lenient (43.9\%), silent (39.5\%), or light responders (38.1\%). Because silent and light responders constitute 65.6\% of the population, the pooled preference falls to 45.2\%; equal weighting across groups raises it to 49.9\% ($p=0.89$). Variation across user characteristics is substantially larger than across task categories: feedback disposition has a spread of 0.248, compared with only 0.040 for the work-life split. The weakest gains appear among light AI users (0.373), rural users (0.402), and light responders (0.381). Genuine profile matching explains 66\% of the effect for expressive users but only 34\% for silent users. These results show that aggregate evaluation can hide important differences in personalization across users and evaluators.

\paragraph{\textbf{Finding 3: Personalization is a selective capability whose value decreases as interaction context becomes informative.}}
Profile conditioning does not simply track overall model quality. GPT-5.5 is the only assistant that benefits from conditioning across all five feedback dispositions, with preference rates ranging from 0.571 for light responders to 0.720 for critical users, despite ranking only sixth in overall turn-level strength. Every other assistant loses in at least three groups, with differences of up to 37.8 percentage points across feedback dispositions. Response length provides one possible explanation: GPT-5.5 is also the only model that does not systematically shorten its responses under profile conditioning, while across assistants, greater shortening is strongly associated with lower preference ($r=0.96$). This suggests that useful personalization requires adapting content rather than applying a generic stylistic transformation.

The value of conditioning also depends on whether the task and dialogue leave room for user-specific adaptation. Across sixteen coarse domains, conditioned responses perform best in current affairs (0.509), travel (0.491), and everyday services (0.485), but worst in law (0.378), shopping (0.379), and computing (0.391). Across task scenarios, creative entertainment is highest (0.502), while professional retrieval is lowest (0.388). Profiles are also substantially more useful at cold start: conditioned responses are preferred at 0.502 without prior dialogue, compared with 0.412 when history is available, declining from 0.427 after one prior turn to 0.415 after two or three and 0.396 after four or more. The same pattern holds across user groups. Together, these results suggest that profiles and conversational history provide partly overlapping information, and that personalization is most useful when the desired response depends on the user, but the interaction itself has not yet revealed sufficient evidence about that user.

\subsubsection{Repairing Real Failures}
\label{sec:res_repair}

\paragraph{\textbf{Finding 4: Assistants repair factual errors more reliably than common interaction failures.}}
Across all benchmark instances, the overall repair rate, defined as the fraction of new responses judged satisfactory by the URM, is 0.429, but performance varies substantially across failure types. Factual errors are repaired most often (0.548), followed by intent mismatch (0.531), whereas verbosity (0.328) and formatting (0.324) are repaired least often. This imbalance is particularly important because verbosity and formatting together account for 34.7\% of the benchmark, compared with only 6.8\% for factual errors. Repair rates increase consistently with overall model strength ($r=0.76$--$0.99$ across failure dimensions), indicating that stronger systems improve broadly rather than specializing in particular failure types. Nevertheless, the most common interaction-level failures remain the hardest to repair, suggesting that progress on factual correctness alone leaves substantial room for improving the experience users encounter in practice.

\subsection{URM-Bench Results}
\label{sec:urm_results}

\subsubsection{Training Strategies}

\begin{table}[t]
\centering
\small
\caption{PairAcc of five trained reward models on the 800-instance URM-Bench under the population condition. Chance is 0.5, exact ties count as errors, and $^{\checkmark}$ marks the model selected as the URM for UXBench Pro.}
\setlength{\tabcolsep}{6pt}
\renewcommand{\arraystretch}{1.08}
\begin{tabular}{l l cccc c}
\toprule
\textbf{Formulation} & \textbf{Training signal} & \textbf{S1$\uparrow$} & \textbf{S2$\uparrow$} & \textbf{S3$\uparrow$} & \textbf{S4$\uparrow$} & \textbf{Pooled$\uparrow$} \\
\midrule
Pointwise GRM & Pointwise feedback labels & 0.5500 & 0.5300 & 0.4650 & 0.5650 & 0.5275 \\
General RM & Mixed, general purpose & 0.6700 & 0.5300 & 0.4850 & 0.6100 & 0.5737 \\
BTRM & Regeneration preference & 0.6900 & 0.6100 & 0.4900 & 0.5900 & 0.5950 \\
BTRM & Online dual-answer choice & 0.6850 & 0.6050 & 0.5400 & 0.6250 & 0.6138 \\
\textbf{Pairwise GRM}$^{\checkmark}$ & Regeneration preference & \textbf{0.8650} & \textbf{0.8050} & \textbf{0.6200} & \textbf{0.7850} & \textbf{0.7688} \\
\bottomrule
\end{tabular}
\label{tab:urm_bench}
\end{table}

\paragraph{\textbf{Finding 5: Reliable UX evaluation requires both an appropriate evaluator and relevant user context.}}
We first examine which evaluator is most reliable for UX judgment. Table~\ref{tab:urm_bench} compares reward-model formulations and training signals on URM-Bench. Under the same Bradley--Terry formulation, dual-answer choice reaches 0.6138 PairAcc and regeneration preference 0.5950, compared with 0.5738 for the general-purpose model and 0.5275 for pointwise feedback. The larger gain comes from the formulation itself: using the same regeneration signal, the pairwise GRM reaches 0.7688, improving over BTRM by 17 points. We therefore use the pairwise GRM trained on regeneration preference as the URM for UXBench Pro.

Having established the evaluator, we next ask whether general-purpose frontier models can provide comparably reliable UX judgments. Table~\ref{tab:urm_profile_dim} shows that they cannot. Without profile conditioning, prompted judges range from 0.5513 to 0.6525 PairAcc, whereas the certified URM reaches 0.7688, outperforming the best prompted judge, GLM-5.1, by 11.6 points and approaching the human reference of 0.8271. The largest gap appears in style adaptation: prompted judges average only 0.536, with eight of eighteen at or below chance, compared with 0.807 for human annotators. Thus, strong general model capability does not directly translate into reliable UX evaluation.

We then examine whether explicit user context can close this gap. Providing the FACTORS profile improves pooled accuracy for 17 of 18 prompted judges and raises the certified URM from 0.7688 to 0.8187, close to the human baseline of 0.8271. The gain is concentrated in style adaptation, where all 19 systems improve by 10.7 points on average and by as much as 16.0 points. In contrast, state recognition and need inference do not improve overall, and long-term preference changes only slightly. This pattern suggests that profile information is most useful when the dialogue alone does not reveal which otherwise plausible response better fits the user.

\begin{table}[t]
\centering
\footnotesize
\caption{Per-mode effect of FACTORS profile conditioning on URM-Bench. Each group reports PairAcc without the profile, with the profile, and their paired difference $\Delta$. The human baseline is shown as a reference for each mode. $^{*}$ indicates exact McNemar $p<0.05$, and $^{**}$ indicates significance after Bonferroni correction.}
\setlength{\tabcolsep}{4.2pt}
\renewcommand{\arraystretch}{1.08}
\adjustbox{max width=\linewidth}{%
\begin{tabular}{r l ccc ccc ccc ccc}
\toprule
& \textbf{Evaluator}
& \multicolumn{3}{c}{\textbf{S1 State}}
& \multicolumn{3}{c}{\textbf{S2 Need}}
& \multicolumn{3}{c}{\textbf{S3 Style}}
& \multicolumn{3}{c}{\textbf{S4 Long-term}} \\
\cmidrule(lr){3-5}
\cmidrule(lr){6-8}
\cmidrule(lr){9-11}
\cmidrule(lr){12-14}
& & \textbf{w/o} & \textbf{w/} & \textbf{$\Delta$}
& \textbf{w/o} & \textbf{w/} & \textbf{$\Delta$}
& \textbf{w/o} & \textbf{w/} & \textbf{$\Delta$}
& \textbf{w/o} & \textbf{w/} & \textbf{$\Delta$} \\
\midrule
1  & \textbf{Pairwise GRM}$^{\checkmark}$ & \textbf{0.865} & \textbf{0.865} & $+0.000$ & \textbf{0.805} & \textbf{0.805} & $+0.000$ & 0.620 & \textbf{0.760} & $\mathbf{+0.140^{**}}$ & \textbf{0.785} & \textbf{0.845} & $+0.060^{*}$ \\
2  & GLM-5.1 & 0.750 & 0.720 & $-0.030$ & 0.645 & 0.620 & $-0.025$ & 0.580 & 0.710 & $+0.130^{*}$ & 0.635 & 0.625 & $-0.010$ \\
3  & Hunyuan-3 & 0.740 & 0.710 & $-0.030$ & 0.695 & 0.615 & $-0.080^{*}$ & 0.570 & 0.680 & $+0.110^{*}$ & 0.585 & 0.645 & $+0.060$ \\
4  & Claude-Opus-4.8 & 0.710 & 0.710 & $+0.000$ & 0.620 & 0.620 & $+0.000$ & 0.590 & 0.680 & $+0.090^{*}$ & 0.650 & 0.625 & $-0.025$ \\
5  & GPT-5.5 & 0.675 & 0.665 & $-0.010$ & 0.580 & 0.610 & $+0.030$ & 0.675 & 0.705 & $+0.030$ & 0.620 & 0.605 & $-0.015$ \\
6  & Claude-Opus-4.7 & 0.725 & 0.675 & $-0.050$ & 0.655 & 0.605 & $-0.050$ & 0.550 & 0.690 & $\mathbf{+0.140^{**}}$ & 0.600 & 0.580 & $-0.020$ \\
7  & GLM-5.2 & \textbf{0.765} & 0.725 & $-0.040$ & 0.615 & 0.565 & $-0.050$ & 0.495 & 0.655 & $\mathbf{+0.160^{**}}$ & 0.635 & 0.615 & $-0.020$ \\
8  & Qwen3.7-Max & 0.680 & 0.720 & $+0.040$ & 0.655 & 0.640 & $-0.015$ & 0.560 & 0.610 & $+0.050$ & 0.595 & 0.640 & $+0.045$ \\
9  & Kimi-K3 & 0.690 & 0.710 & $+0.020$ & 0.665 & 0.625 & $-0.040$ & 0.525 & 0.685 & $\mathbf{+0.160^{**}}$ & 0.590 & 0.665 & $+0.075^{*}$ \\
10 & GPT-5.4 & 0.690 & 0.680 & $-0.010$ & 0.560 & 0.565 & $+0.005$ & 0.595 & 0.685 & $+0.090^{*}$ & 0.610 & 0.595 & $-0.015$ \\
11 & Claude-Sonnet-4.6 & 0.715 & 0.680 & $-0.035$ & 0.625 & 0.510 & $-0.115^{*}$ & 0.525 & 0.675 & $\mathbf{+0.150^{**}}$ & 0.585 & 0.580 & $-0.005$ \\
12 & Doubao-Seed-2.0-Pro & 0.720 & 0.680 & $-0.040$ & 0.645 & 0.610 & $-0.035$ & 0.500 & 0.625 & $+0.125^{*}$ & 0.565 & 0.550 & $-0.015$ \\
13 & Gemini-3.1-pro & 0.680 & 0.710 & $+0.030$ & 0.645 & 0.595 & $-0.050$ & 0.500 & 0.585 & $+0.085^{*}$ & 0.585 & 0.600 & $+0.015$ \\
14 & Gemini-3-flash & 0.660 & 0.675 & $+0.015$ & 0.630 & 0.610 & $-0.020$ & 0.490 & 0.580 & $+0.090^{*}$ & 0.610 & 0.615 & $+0.005$ \\
15 & GLM-5 & 0.705 & 0.720 & $+0.015$ & 0.625 & 0.620 & $-0.005$ & 0.500 & 0.640 & $\mathbf{+0.140^{**}}$ & 0.540 & 0.575 & $+0.035$ \\
16 & DeepSeek-v4-pro & 0.655 & 0.615 & $-0.040$ & 0.670 & 0.685 & $+0.015$ & 0.475 & 0.525 & $+0.050$ & 0.550 & 0.550 & $+0.000$ \\
17 & DeepSeek-v4-flash & 0.675 & 0.670 & $-0.005$ & 0.635 & 0.650 & $+0.015$ & 0.460 & 0.610 & $\mathbf{+0.150^{**}}$ & 0.560 & 0.630 & $+0.070$ \\
18 & Gemini-2.5-flash & 0.640 & 0.635 & $-0.005$ & 0.630 & 0.655 & $+0.025$ & 0.460 & 0.545 & $+0.085^{*}$ & 0.580 & 0.605 & $+0.025$ \\
19 & GPT-5.2 & 0.620 & 0.670 & $+0.050$ & 0.475 & 0.500 & $+0.025$ & 0.590 & 0.655 & $+0.065^{*}$ & 0.520 & 0.580 & $+0.060$ \\
\midrule
\multicolumn{2}{l}{\textit{Mean $\Delta$, systems improving}}
& \multicolumn{3}{c}{\textit{$-0.007$, 6/19}}
& \multicolumn{3}{c}{\textit{$-0.019$, 6/19}}
& \multicolumn{3}{c}{\textbf{\textit{$+0.107$, 19/19}}}
& \multicolumn{3}{c}{\textit{$+0.017$, 10/19}} \\
\midrule
\multicolumn{2}{l}{\textit{Human baseline}}
& \multicolumn{3}{c}{0.870}
& \multicolumn{3}{c}{0.802}
& \multicolumn{3}{c}{0.807}
& \multicolumn{3}{c}{0.829} \\
\bottomrule
\end{tabular}%
}
\label{tab:urm_profile_dim}
\end{table}

Two additional analyses test whether this improvement genuinely comes from user information rather than incidental prompting effects. First, perturbing only the profile content shows that the URM responds to profile semantics rather than to the mere presence of profile text. A semantically empty placebo changes pooled accuracy by only 0.1 points ($p=1.000$), whereas reversing all seven facets changes accuracy by 7.26 points. The largest effect again appears in style adaptation, where joint reversal shifts accuracy by 17.36 points ($p<10^{-4}$), and changing expertise alone shifts it by 10.96 points, while geography has essentially no effect. A stricter congruence test on technical proficiency is not significant (258 pairs, $-2.3$ points, $p=0.34$), indicating that the URM is sensitive to profile content but does not necessarily interpret every facet in a fully calibrated way.

Second, we test whether additional inference-time computation can substitute for this missing user context. It cannot. In the $2\times2$ ablation in Table~\ref{tab:urm_thinking}, higher thinking effort helps only DeepSeek-v4-pro, while Hunyuan-3, GPT-5.5, and Gemini-3.1-pro perform worse. The only significant changes are declines: GPT-5.5 drops by 5.1 points without a profile ($p=0.002$), and Hunyuan-3 drops by 4.1 points with one ($p=0.036$). We therefore use the non-deliberative setting in the main evaluation. Overall, these results show that reliable UX judgment depends on choosing an evaluator suited to preference comparison and providing user context where preference is genuinely user-dependent; stronger generic models or additional deliberation do not reliably recover either requirement.

\begin{table}[t]
\centering
\small
\caption{Thinking-effort ablation on four judges. Values are pooled PairAcc under low and high thinking effort, with and without profile conditioning. The best result for each model is bolded.}
\setlength{\tabcolsep}{7pt}
\renewcommand{\arraystretch}{1.08}
\begin{tabular}{l cc cc}
\toprule
& \multicolumn{2}{c}{\textbf{Low effort}} & \multicolumn{2}{c}{\textbf{High effort}} \\
\cmidrule(lr){2-3}\cmidrule(lr){4-5}
\textbf{System} & \textbf{Base} & \textbf{+Profile} & \textbf{Base} & \textbf{+Profile} \\
\midrule
DeepSeek-v4-pro & 0.5875 & 0.5938 & 0.6175 & \textbf{0.6400} \\
Hunyuan-3 & 0.6475 & \textbf{0.6625} & 0.6150 & 0.6212 \\
GPT-5.5 & 0.6375 & \textbf{0.6462} & 0.5863 & 0.6175 \\
Gemini-3.1-pro & 0.6025 & \textbf{0.6225} & 0.5913 & 0.5887 \\
\bottomrule
\end{tabular}
\label{tab:urm_thinking}
\end{table}

\subsection{User Simulator}
\label{sec:res_usim}

\subsubsection{Linguistic and Behavioral Fidelity}

\paragraph{\textbf{Finding 6: Retrieval improves linguistic fidelity, but realistic user behavior remains strongly dependent on the base model.}}
Across the twelve base models in Table~\ref{tab:usim_rag}, retrieval improves dialogue-level semantic similarity and turn-aligned similarity for 11 of 12 models and reduces aggregate divergence for 9 of 12. This consistent gain shows that a profile alone does not specify how a user actually communicates: retrieved utterances provide concrete examples of realistic phrasing, length, and conversational reactions. The improvement, however, does not extend consistently to feedback behavior. Among ten models with complete signal results, retrieval improves explicit-signal Macro F1, implicit-signal Macro MCC, and turn-level MCC for only three models each, with average changes of $-0.0250$, $-0.0297$, and $-0.0095$, respectively. Linguistic resemblance is therefore insufficient evidence of behavioral fidelity.

\begin{table}[t]
\centering 
\footnotesize 
\caption{Effect of retrieval augmentation across twelve base models on USimBench.}
\setlength{\tabcolsep}{4.5pt}
\renewcommand{\arraystretch}{1.08}
\adjustbox{max width=\linewidth}{%
\begin{tabular}{r l ccc ccc ccc}
\toprule
& \textbf{Base model}
& \multicolumn{3}{c}{\textbf{Dialogue cosine$\uparrow$}}
& \multicolumn{3}{c}{\textbf{Turn-aligned area$\uparrow$}}
& \multicolumn{3}{c}{\textbf{Aggregate divergence$\downarrow$}} \\
\cmidrule(lr){3-5}\cmidrule(lr){6-8}\cmidrule(lr){9-11}
& & \textbf{Profile-only} & \textbf{RAG} & \textbf{$\Delta$}
& \textbf{Profile-only} & \textbf{RAG} & \textbf{$\Delta$}
& \textbf{Profile-only} & \textbf{RAG} & \textbf{$\Delta$} \\
\midrule
1  & Kimi-K3 & 0.5904 & 0.6285 & $+0.0381$ & 0.4831 & \textbf{0.5383} & $+0.0552$ & 0.1993 & \textbf{0.1697} & $-0.0296$ \\
2  & Hunyuan-3 & 0.5875 & \textbf{0.6331} & $+0.0456$ & 0.4671 & 0.5068 & $+0.0397$ & 0.1964 & 0.1712 & $-0.0252$ \\
3  & Claude-Opus-4.8 & 0.5769 & 0.6137 & $+0.0368$ & 0.4601 & 0.4919 & $+0.0318$ & 0.1958 & 0.1760 & $-0.0198$ \\
4  & Gemini-3-Flash & 0.6013 & 0.6087 & $+0.0074$ & 0.4977 & 0.5246 & $+0.0269$ & 0.2086 & 0.1777 & \textbf{$-0.0309$} \\
5  & GLM-5.2 & 0.5767 & 0.5885 & $+0.0118$ & 0.4748 & 0.5034 & $+0.0286$ & 0.1930 & 0.1808 & $-0.0122$ \\
6  & GPT-5.5 & 0.6109 & 0.6216 & $+0.0107$ & 0.5031 & 0.5075 & $+0.0044$ & 0.2044 & 0.1833 & $-0.0211$ \\
7  & Doubao-Seed-2.0-Pro & 0.5698 & 0.5731 & $+0.0033$ & 0.4588 & 0.4686 & $+0.0098$ & 0.2001 & 0.1843 & $-0.0158$ \\
8  & Deepseek-v4-Pro & 0.5731 & 0.6084 & $+0.0353$ & 0.4880 & 0.5187 & $+0.0307$ & 0.1963 & 0.1876 & $-0.0087$ \\
9  & Deepseek-v4-Flash & 0.6068 & 0.6140 & $+0.0072$ & 0.4920 & 0.5002 & $+0.0082$ & 0.1950 & 0.1979 & $+0.0029$ \\
10 & Qwen3.7-Max & 0.5741 & 0.5870 & $+0.0129$ & 0.4852 & 0.4841 & $-0.0011$ & 0.1995 & 0.2009 & $+0.0014$ \\
11 & Gemini-3.1-Pro & 0.5696 & 0.4553 & $-0.1143$ & 0.5027 & 0.5375 & $+0.0348$ & \textbf{0.1880} & 0.2121 & $+0.0241$ \\
12 & GPT-5.2 & \textbf{0.6127} & 0.6314 & $+0.0187$ & \textbf{0.5118} & 0.5210 & $+0.0092$ & 0.2317 & 0.2216 & $-0.0101$ \\
\bottomrule
\end{tabular}%
}
\label{tab:usim_rag}
\end{table}

Substantial variation also remains across base models under the same simulation method. Hunyuan-3 achieves the highest style similarity both with and without retrieval (0.8572 and 0.8569), whereas GPT-5.2 is the lowest in both settings (0.5712 and 0.5604). Repetition produces a different ranking: GPT-5.2 exceeds real-user repetition rates by $+0.1862$ and $+0.1837$, while Claude-Opus-4.8 is closest to authentic repetition behavior ($+0.0092$ and $-0.0014$) despite only moderate style similarity. Thus, neither semantic similarity nor style matching alone is sufficient for selecting a user simulator; behavioral characteristics of the underlying model remain consequential even after retrieval augmentation.

\subsubsection{Dynamic State and Profile Effects}

\paragraph{\textbf{Finding 7: Faithful first-person simulation requires dynamic state representations and produces observable profile-dependent behavior.}}
We first examine whether the simulator's reported states actually evolve with the interaction. Table~\ref{tab:state_collapse} shows that generic five-point scales often produce persistent states: across base models, all four state coordinates remain unchanged after their first value in 8.2\% to 82.0\% of multi-turn sessions, with Hunyuan-3 showing the strongest collapse. Using behaviorally anchored descriptions and requiring evidence before each rating substantially reduces this persistence, with joint collapse remaining at most 5.7\% among the reported anchored models and reaching zero for GPT-5.5 and Claude-Opus-4.8. Because the two prompting schemes are evaluated on different base models, this comparison is diagnostic rather than a controlled ablation, but it highlights state persistence as an important failure mode when LLM-generated ratings are used to represent evolving experience.

Beyond being dynamic, the reported states also move coherently with the simulator's observable reactions. Across 7,333 scored turns, all four dimensions decrease when the simulator emits a negative signal, such as dislike, regenerate, or stop. Goal progress shows the largest drop, from 4.048 to 2.319, followed by emotional fit and engagement preference, while cognitive fit changes less, by 0.856 points. This pattern is consistent with the intended distinction between task progress and presentation: an interaction can remain easy to process while still failing to advance the user's goal or meet their expectations.

We finally ask whether the injected profile changes the simulator's behavior rather than only its self-reported states. It does. Silent users emit an explicit signal in only 7.8\% of sessions, compared with 37.7\% for light-feedback users, 68.1\% for critical users, and more than 97\% for lenient and expressive users. The same ordering appears at the turn level, ranging from 2.5\% for silent users to nearly 90\% for the most expressive groups. Together, these results show that faithful first-person simulation requires both responsive state representations and profile-dependent behavioral variation. At the same time, consistency between simulated states and simulated behavior should not be interpreted as external validation of the latent states themselves; USimBench primarily evaluates fidelity with respect to observable user continuations and behavior.

\begin{table}[t]
\centering 
\small
\caption{State persistence in multi-turn RAG user simulation. Each value is the fraction of sessions in which a state coordinate, or all four jointly, remains unchanged after the first turn. $n$ denotes the number of sessions with at least two scored turns.}
\setlength{\tabcolsep}{6pt}
\renewcommand{\arraystretch}{1.08}
\begin{tabular}{l r cccc c}
\toprule
\textbf{Base model} & \textbf{$n$} & \textbf{S1$\downarrow$} & \textbf{S2$\downarrow$} & \textbf{S3$\downarrow$} & \textbf{S4$\downarrow$} & \textbf{All four$\downarrow$} \\
\midrule
\multicolumn{7}{l}{\textit{Generic scale}} \\
Deepseek-v4-Pro   & 117 & 0.248 & 0.504 & 0.368 & 0.376 & 0.120 \\
Deepseek-v4-Flash & 147 & 0.197 & 0.327 & 0.367 & 0.286 & 0.082 \\
GPT-5.2           & 150 & 0.307 & 0.573 & 0.553 & 0.540 & 0.260 \\
Hunyuan-3         & 150 & 0.833 & 0.947 & 0.987 & 0.953 & 0.820 \\
\midrule
\multicolumn{7}{l}{\textit{Anchored variant}} \\
GPT-5.5              & 142 & \textbf{0.007} & 0.268 & \textbf{0.070} & \textbf{0.127} & \textbf{0.000} \\
Claude-Opus-4.8      & 141 & 0.113 & 0.277 & 0.085 & 0.149 & \textbf{0.000} \\
GLM-5.2              & 106 & 0.208 & 0.226 & 0.170 & 0.245 & 0.009 \\
Qwen3.7-Max          & 112 & 0.179 & \textbf{0.143} & 0.134 & 0.152 & 0.027 \\
Kimi-K3              & 121 & 0.215 & 0.339 & 0.165 & 0.364 & 0.050 \\
Doubao-Seed-2.0-Pro  & 123 & 0.171 & 0.537 & 0.390 & 0.293 & 0.057 \\
\bottomrule
\end{tabular}
\label{tab:state_collapse}
\end{table}

\subsection{Discussion: Implications for User-Centric Evaluation}
\label{sec:res_discussion}

Taken together, the results suggest that personalized conversational evaluation cannot be reduced to either a stronger response judge or a more detailed user profile. Profile conditioning changes evaluation most when the preferred response genuinely depends on the user and when conversational evidence is sparse, but its effect can disappear under population averaging or a profile-blind evaluator. Personalization capability is also distinct from general model quality: stronger assistants do not uniformly exploit profile information better, and generic adaptation, such as shortening responses, can reduce rather than improve preference. Evaluation should therefore report subgroup-level effects and distinguish situations in which the user context is actionable from those in which the task is largely user-independent.

Multi-turn evaluation reveals a complementary limitation of response-level benchmarking. Resolution, abandonment, and pairwise preference are related but capture different properties of the interaction, while goal progress appears more closely associated with abandonment than cognitive fit. Similarly, user simulators that reproduce surface language more faithfully do not necessarily reproduce feedback behavior, and static rating scales can fail to reflect changes across turns. These results motivate treating conversational UX as a collection of behaviorally grounded signals rather than collapsing it into a single universal score. More broadly, they reinforce the role of meta-benchmarking: before computational evaluators are used to compare assistants, their sensitivity to user context and their fidelity to observed human preferences and behavior should themselves be established.

\section{Conclusion and Future Work}
\label{sec:conclusion}

This paper introduces UXBench Pro, a personalized benchmark for user experience evaluation that treats satisfaction as jointly shaped by the user and the task rather than as a single population-level judgment. Our results show that this distinction matters in practice: the apparent value of personalization depends on both whose experience is being evaluated and whether the evaluator has access to the corresponding user profile. Profile conditioning is most useful when conversational context is sparse, and the preferred response genuinely depends on the user, while generic evaluators can obscure these effects through population averaging or profile mismatch. We further show that first-person simulation complements response-level ranking by exposing interaction dynamics such as goal progress, abandonment, and evolving cognitive state dimensions. Taken together, UXBench Pro, Sim4Eval, URM-Bench, and USimBench support a broader HCI view of conversational AI evaluation, in which system quality is assessed not only by whether a response is correct, but also by how well it fits a particular user and how that experience evolves over the course of interaction.

Several directions follow from these findings. Future work should evaluate frontier assistants across user subgroups to identify who benefits from model progress and who remains underserved, and should develop profile representations that capture actionable goals, knowledge, and preferences rather than only static attributes. Reward models could be trained with objectives that jointly capture relative preference, absolute satisfaction, and interpretable UX dimensions such as trust, progress, and interaction risk. More adaptive evaluators could generate task-specific rubrics, verify factual claims with tools, and track satisfaction turn by turn before distilling these judgments into efficient models. Finally, FACTORS should be extended beyond a single platform through longer-term preference memory, richer behavioral signals, and cross-lingual and cross-cultural validation. From an HCI perspective, the central challenge is not simply to make assistants more personalized, but to design evaluation methods that can identify when personalization is useful, for whom it works, and how it changes the course of an interaction.

\section*{Ethical Considerations}

All data originates from an opt-in experience program of the AI assistant, in which users voluntarily enroll and explicitly authorize the use of their interaction logs for research and service improvement, so that no session enters the pipeline without consent obtained before collection. A locally deployed LLM fully rewrites the data to further reduce the risk of retaining identifying information, and is further protected by strict data cleaning and de-identification filtering, as mentioned in the construction pipeline. FACTORS dimensions are coarse contextual proxies rather than definitive user attributes and may encode demographic or socioeconomic biases; subgroup differences should therefore not be interpreted causally or used for consequential individual-level decisions. The released data should be used under privacy-preserving protocols and only for model evaluation, not for model training.

\bibliography{references}
\bibliographystyle{ACM-Reference-Format}

\clearpage
\appendix

\section{\emph{FACTORS} Profile Encoding}
\label{appendix:profile_encoding}

The \emph{\textbf{FACTORS}} schema encodes each user along seven behavioral facets derived from logged
interaction signals rather than from self-reported survey data. Each facet is assigned from a fixed closed vocabulary; the
special suffix~\textsf{X} marks any facet whose source signals are absent or contradictory.

\begin{description}[leftmargin=2.2em, labelwidth=1.8em, labelsep=0.4em,
                  font=\normalfont\bfseries]

\item[F -- Feedback disposition (core axis).]
Derived from the ratio of positive (like, copy, share) to negative (unlike, regenerate) interactions over the preceding
30~days, normalized by total prompt count.
\begin{itemize}[nosep, leftmargin=1.6em]
\item \textsf{F1}~Silent: total interaction count is zero, or the interaction-to-prompt ratio is below 1\%.
\item \textsf{F2}~Critical: at least two negative signals and fewer than 20\% of interactions are positive.
\item \textsf{F3}~Lenient: at least two positive signals and more than 70\% of interactions are positive.
\item \textsf{F4}~Expressive: at least five total signals with a positive fraction between 20\% and 70\% (balanced but
vocal).
\item \textsf{F5}~Light: residual category; occasional signals that do not meet any threshold above.
\end{itemize}

\item[A -- Age.]
Derived from the platform-recorded age interval associated with the user account.
\begin{itemize}[nosep, leftmargin=1.6em]
\item \textsf{A1}: 18 years or younger.
\item \textsf{A2}: 19--30 years.
\item \textsf{A3}: 31--50 years.
\item \textsf{A4}: 51 years or older.
\end{itemize}

\item[C -- Consumption capacity.]
Inferred from the device's market price tier (\emph{phone\_market\_price\_cate}), which is the primary signal because the
platform wealth label disagrees with the actual device price for roughly 20\% of users. Strong asset signals (house ownership,
luxury-housing group membership, employment at a prominent company) override the device tier upward to the highest level.
\begin{itemize}[nosep, leftmargin=1.6em]
\item \textsf{C0}~Low: entry-level handset (budget tier).
\item \textsf{C1}~Mid: mid-range handset.
\item \textsf{C2}~High: high-end handset; also assigned when the platform consumption label is ``high'' and the device
price is missing.
\item \textsf{C3}~Flagship: flagship handset, or any strong asset signal present.
\end{itemize}

\item[T -- Technology usage intensity.]
Derived from recent AI usage signals, including the 30-day sent-prompt count
(\emph{sent\_prompt\_cnt\_30d}) and whether a competing AI product
(DeepSeek or Doubao) is installed. Thresholds are calibrated on the full user
base (Q1~=~19, Q3~=~218).
\begin{itemize}[nosep, leftmargin=1.6em]
\item \textsf{T0}~Light: fewer than 19 prompts and no competing product installed.
\item \textsf{T1}~Regular: 19 or more prompts, or any competing product installed.
\item \textsf{T2}~Heavy: 218 or more prompts, or a competing product installed and at least 75 prompts.
\end{itemize}

\item[O -- Occupation role.]
Inferred from dialogue-mined and platform life-stage signals, including age,
college-student status, and parenting status. Signals are applied in priority
order to avoid contradictory assignments.
\begin{itemize}[nosep, leftmargin=1.6em]
\item \textsf{O1}~Student: college-student flag set, or age is \textsf{A1} and the parenting flag is absent.
\item \textsf{O2}~Working adult: age is \textsf{A2} or \textsf{A3} and neither the parenting flag nor the college-student flag is set.
\item \textsf{O3}~Retired/senior: age is \textsf{A4}.
\item \textsf{O4}~Parent with young children: parenting stage field is populated.
\end{itemize}

\item[R -- Region.]
Derived from the 30-day residential city name (\emph{residence\_city\_name}), with the rural-group flag
(\emph{is\_rural\_group}) as an override. Tier boundaries follow the State Council 2014 city-size classification and the
Ministry of Housing and Urban-Rural Development 2023 Statistical Yearbook.
\begin{itemize}[nosep, leftmargin=1.6em]
\item \textsf{R1}~Metropolis: direct-controlled municipalities, provincial capitals and autonomous-region capitals,
sub-provincial cities, special administrative regions, and cities with an urban resident population of at least five
million.
\item \textsf{R2}~Town: any other city with a municipal designation (name ends in \begin{CJK*}{UTF8}{gbsn}市\end{CJK*}).
\item \textsf{R3}~Rural: rural-group flag set, or the city name ends in a county-level administrative suffix
(\begin{CJK*}{UTF8}{gbsn}县、旗、盟、地区、自治州\end{CJK*}).
\end{itemize}

\item[S -- Schooling.]
Derived from platform education signals and mapped into three coarse schooling levels.
\begin{itemize}[nosep, leftmargin=1.6em]
\item \textsf{S0}~Primary/lower secondary: primary school or middle school.
\item \textsf{S1}~Upper secondary: high school or vocational college
(\begin{CJK*}{UTF8}{gbsn}大专\end{CJK*}).
\item \textsf{S2}~Tertiary and above: bachelor's degree or higher, including master's and doctoral degrees.
\end{itemize}

\end{description}

\section{Human Annotators and Ethical Considerations}
\label{appendix:annotators}

Ten recruited employees annotated as part of their regular duties during normal working hours, with no expectation of overtime or uncompensated participation. Training included task-specific guidelines, worked examples, and initial calibration; briefings covered the task purpose, potential content risks, and escalation procedures. Data were rewritten, safety-screened, and de-identified by rule-based systems before review. Annotators saw only task-relevant content, with metadata and user profiles withheld when unnecessary for judgment, and were instructed not to infer or record information beyond the annotation schema. They were encouraged to escalate uncertainty and could skip or escalate sensitive cases without penalty. Quality control combined independent double annotation, adjudication, agreement monitoring, and periodic calibration.

\section{Task Taxonomy and Scenario Assignment}
\label{appendix:task_assignment}

Each trace is assigned to one of the 12 scenarios in Table~\ref{tab:task_rules} with Hunyuan-3 as the annotator. Rules are evaluated from top to bottom, with high-confidence intent fields taking precedence over industry labels, and unmatched cases are assigned to LF-g. Each session receives the modal trace-level scenario.

\begin{table}[t]
\centering 
\small
\caption{Deterministic trace-level scenario assignment in UXBench Pro.}
\setlength{\tabcolsep}{4pt}
\begin{tabular}{@{}p{0.09\linewidth}p{0.27\linewidth}p{0.57\linewidth}@{}}
\toprule
\textbf{Code} & \textbf{Scenario} & \textbf{Assignment signal} \\
\midrule
WK-a & Office writing & Workplace, business, or professional writing intent \\
WK-b & Programming & Code intent or programming meta-intent \\
WK-c & Analysis and decision & Analysis or decision intent; finance for working-role users \\
WK-d & Professional retrieval & Legal industry \\
WK-e & Business translation & Translation intent \\
LF-a & Learning assistance & Mathematics intent, education, or scholastic knowledge \\
LF-b & Everyday services & Shopping, public services, travel, or current affairs \\
LF-c & Health and wellness & Medical or health industry \\
LF-d & Emotional support and chat & Chat intent or entertainment industry \\
LF-e & Creative entertainment & Creative writing, image generation, or image understanding intent \\
LF-f & Personal finance & Finance industry for users not assigned the working role \\
LF-g & General information & No preceding rule applies \\
\bottomrule
\end{tabular}
\label{tab:task_rules}
\end{table}

\section{User-State Scoring Rubric}
\label{appendix:state_rubric}

Each dimension is rated independently from the simulated user's first-person perspective on an integer scale from 1 to 5. The prompt requests brief evidence before each score; unparsable values are recorded as missing.

\begin{description}[leftmargin=2.4em,labelwidth=2.0em,labelsep=0.4em,font=\normalfont\bfseries]
\item[S1 Goal progress.]
1: off topic; 2: relevant but unusable; 3: partial progress with a key step missing; 4: essentially complete with minor gaps; 5: complete without a necessary follow-up.

\item[S2 Emotional fit.]
1: distinctly uncomfortable or mismatched; 2: polite but misses the user's emotional state; 3: neutral; 4: tone fits the expressed state; 5: also recognizes an unstated concern, urgency, or hesitation.

\item[S3 Cognitive fit.]
1: unreadable due to length, jargon, or disorder; 2: requires substantial rereading or skipping; 3: understandable but over- or under-detailed; 4: mostly appropriate length and depth; 5: immediately usable and appropriately complete. This dimension assesses presentation, not factual correctness.

\item[S4 Engagement preference.]
1: strongly violates the desired interaction mode; 2: somewhat over- or under-interactive; 3: acceptable default interaction; 4: asks or acts at an appropriate level; 5: anticipates exactly the next interaction the user would want.
\end{description}

\section{Experiment Configuration and Statistical Analysis}
\label{appendix:statistics}

\paragraph{Generation and judging.}
Default assistant generation uses a common conversation template, temperature zero, reasoning disabled, an 8,192-token output limit, and no external tools. Profile conditions differ only in whether the deterministic persona card is inserted. Prompted URM-Bench judges use temperature zero, reasoning disabled, and a 16-token output limit. The certified pairwise GRM compares both responses in a single forward pass and normalizes the positive and negative judgment-token probabilities to obtain a preference probability.

\paragraph{URM-Bench.}
Pairwise accuracy is
\begin{equation}
\operatorname{PairAcc}
=
\frac{1}{N}\sum_{i=1}^{N}\mathbbm{1}[s_i^{+}>s_i^{-}],
\end{equation}
where $s_i^{+}$ and $s_i^{-}$ score the behaviorally preferred and alternative responses. Exact ties, missing outputs, and unparsable verdicts receive zero credit and remain in the denominator. Profile effects are tested using an exact two-sided McNemar test. The human reference is the mean PairAcc across ten contracted annotators, reported separately from that of automated judges.

\paragraph{Simulation outcomes and fidelity.}
For Sim4Eval, resolution is the fraction of trajectories reaching S1$=5$, and time to resolution is the first such turn. Abandonment requires an early simulator stop with final S1$\leq2$. State means and proportions of scores $\geq4$ use only valid parsed states. USimBench reports feedback-signal fidelity, semantic similarity, and behavioral similarity separately rather than averaging them into a single score, since they assess distinct aspects of simulator fidelity.

\end{document}